\documentclass[11pt,a4paper]{article}
\usepackage[hyperref]{eacl2021}
\usepackage{times}
\usepackage{latexsym}
\usepackage{graphicx}
\usepackage{booktabs}
\usepackage{stfloats}

\usepackage{array}
\newcolumntype{x}[1]{>{\centering\arraybackslash\hspace{0pt}}p{#1}}

\usepackage{microtype}

\aclfinalcopy % Uncomment this line for the final submission
\title{{L}everaging Passage Retrieval with Generative Models \\ for Open Domain Question Answering}

\author{
  Gautier Izacard$^{1,2,3}$ \hspace{2em} Edouard Grave$^1$ \\
  $^1$ Facebook AI Research, Paris \\
  $^2$ ENS, PSL University, Paris \\
  $^3$ Inria, Paris \\
  \texttt{gizacard|egrave@fb.com} \\
}

\date{}

\begin{document}
 \maketitle
\begin{abstract}
Generative models for open domain question answering have proven to be competitive, without resorting to external knowledge.
While promising, this approach requires to use models with billions of parameters, which are expensive to train and query.
In this paper, we investigate how much these models can benefit from retrieving text passages, potentially containing evidence.
We obtain state-of-the-art results on the Natural Questions and TriviaQA open benchmarks.
Interestingly, we observe that the performance of this method significantly improves when increasing the number of retrieved passages.
This is evidence that sequence-to-sequence models offers a flexible framework to efficiently aggregate and combine evidence from multiple passages.
%This is evidence that generative models are good at aggregating and combining evidence from multiple passages.
\end{abstract}

\section{Introduction}
Recently, several works have shown that factual information can be extracted from large scale language models trained on vast quantities of data~\citep{radford2019language,petroni2019language,jiang2019can,talmor2019olmpics}.
Building on that observation and the advances in pretraining of natural language processing models, \citet{roberts2020much} introduced a generative model for open domain question answering.
Without relying on external knowledge, this method obtained competitive results on several benchmarks.
However, it requires models containing billions of parameters, since all the information needs to be stored in the weights.
This makes models expensive to query and train.
In this paper, we investigate how much this method could benefit from having access to an external source of knowledge, such as Wikipedia.

\begin{figure}[t]
\begin{center} 
\includegraphics[width=0.45 \textwidth]{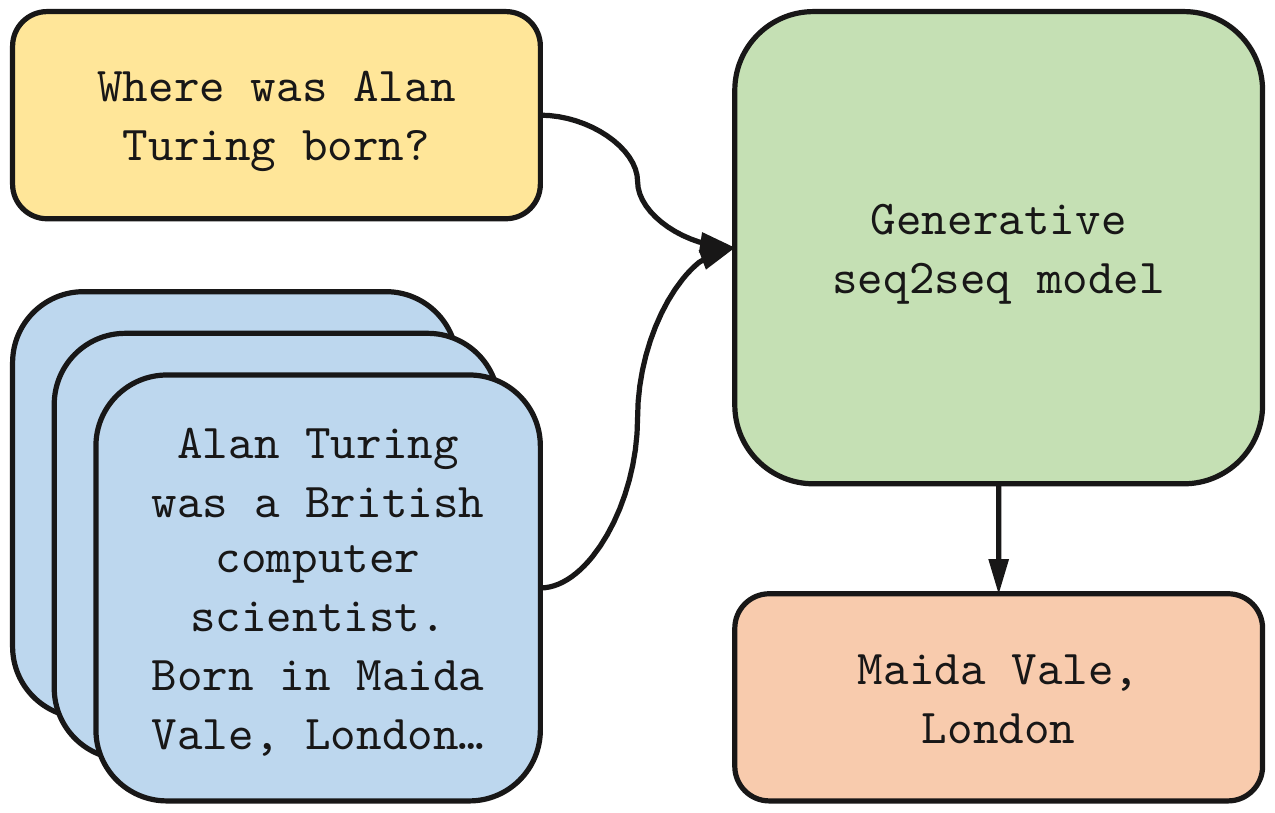}
\vspace{1em}

\caption{A simple approach to open domain question answering.
First, it retrieves support text passages from an external source of knowledge such as Wikipedia.
Then, a generative encoder-decoder model produces the answer, conditioned on the question and the retrieved passages.
%One of our main findings is that this approach scales well with the number of retrieved passages, as the performance keeps improving when retrieving one hundred passages.
This approach scales well with the number of retrieved passages, as the performance keeps improving when retrieving up to one hundred passages.
}
\label{fig:pullfigure}
\end{center}
\end{figure}

\begin{figure*}[h]
\begin{center} 
\includegraphics[width= 0.98 \textwidth]{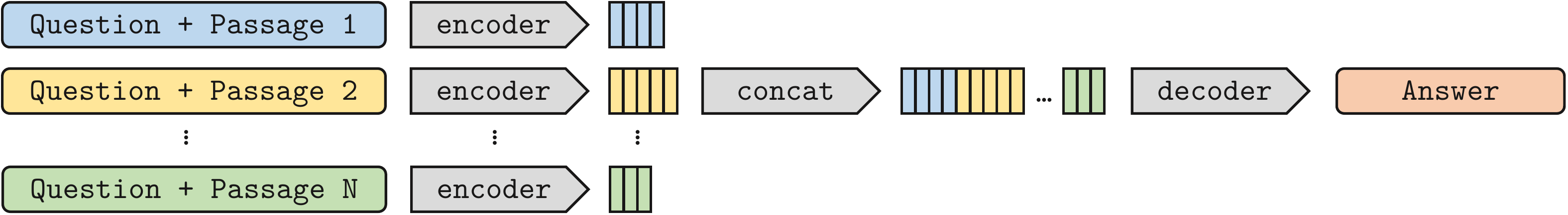}

\caption{Architecture of the Fusion-in-Decoder method.}
\label{fig:FID}
\end{center}
\end{figure*}

Retrieval based approaches were previously considered in the context of open domain question answering with extractive models~\citep{chen2017reading}.
In that case, systems start by retrieving support documents, before extracting the answer from these documents.
Different retrieval techniques have been considered, either using sparse representations based on TF/IDF or using dense embeddings~\citep{guu2020realm,karpukhin2020dense}.
The models which extract the answers are often based on contextualized word representations such as ELMo or BERT~\citep{peters2018deep,devlin2018bert}, and predict a span as answer.
Aggregating and combining evidence from multiple passages is not straightforward when using extractive models, and multiple techniques have been proposed to address this limitation~\citep{clark2017simple,min2019discrete}.

In this paper, we explore a simple approach having the best of both worlds, by building on the exciting developments in generative modeling and retrieval for open domain question answering.
This method proceeds in two steps, by first retrieving supporting passages using either sparse or dense representations.
Then, a sequence-to-sequence model generates the answer, taking as input the retrieved passages in addition to the question.
While conceptually simple, this method sets new state-of-the-art results on the TriviaQA and NaturalQuestions benchmarks.
In particular, we show that the performance of our method significantly improves when the number of retrieved passages increases.
We believe that this is evidence that generative models are good at combining evidence from multiple passages, compared to extractive ones.

\section{Related work}
\paragraph{Open domain question answering} is the task of answering general domain questions, in which the evidence is not given as input to the system.
While being a longstanding problem in natural language processing~\citep{voorhees1999trec}, this task has recently regained interest following the work by \citet{chen2017reading}.
In that version of the problem, strong supervision is available to the learning system, in the form of spans corresponding to answers.
\citet{chen2017reading} proposed to solve the problem by first retrieving support document from Wikipedia, before extracting the answer from the retrieved document.
Different methods were proposed to tackle the setting where no gold spans are given to the system, but only the correct answer.
\citet{clark2017simple} proposed to use a global normalization over all the span corresponding to the answer, which was later applied to BERT based models~\citep{wang2019multi}.
\citet{min2019discrete} introduced a method based on hard expectation-maximization to tackle noisy supervision from this setting.
\citet{wang2017evidence} described a technique to aggregate answers from different paragraphs, using confidence and coverage scores.

\paragraph{Passage retrieval} is an important step in open domain question answering, and is an active area of research to improve QA systems.
Initially, sparse representations based on TF/IDF were used to retrieve support documents~\citep{chen2017reading}.
\citet{lee2018ranking} introduced a supervised learning method to rerank paragraphs based on BiLSTM, while \citet{wang2018r} trained a ranking system with reinforcement learning.
A second approach to improve the retrieval step of QA systems is to used additional information such as the Wikipedia or Wikidata graphs~\citep{min2019knowledge,asai2019learning}.
Recently, multiple works show that retrieval systems entirely based on dense representation and approximate nearest neighbors were competitive with traditional approaches.
Such models can be trained using weak supervision in the form of question-answer pairs~\citep{karpukhin2020dense}, or pretrained using a cloze task and finetuned end-to-end~\citep{guu2020realm,lee2019latent}.

\paragraph{Generative question answering} was mostly considered in previous work for datasets requiring to generate answers, such as NarrativeQA~\citep{kovcisky2018narrativeqa}, CoQA~\citep{reddy2019coqa} or ELI5~\citep{fan2019eli5}.
These datasets were generated in a way that answers do not correspond to spans in support documents, thus requiring abstractive models.
\citet{raffel2019exploring} showed that generative models are competitive for reading comprehension tasks such as SQuAD~\citep{rajpurkar2016squad}, where answers are spans.
\citet{roberts2020much} proposed to use large pretrained generative models, without using additional knowledge, for open domain question answering.
Closest to our work, \citet{min2020ambigqa} and \citet{lewis2020retrieval} introduced retrieval augmented generative models for open domain question answering.
Our approach differs from these works by how the generative model processes the retrieved passages.
This allows to scale to large numbers of documents, and to benefit from this large amount of evidence.

\section{Method}
In this section, we describe our approach to open domain question answering.
It proceeds in two steps, first retrieving support passages before processing them with a sequence to sequence model.

\begin{table*}[t]
\centering
\begin{tabular}{lx{1.2cm}x{1.2cm}x{1.2cm}x{1.2cm}x{1.2cm}}
  \toprule
  Model & NQ & \multicolumn{2}{c}{TriviaQA} & \multicolumn{2}{c}{SQuAD Open} \\
  & EM & EM & EM & EM & F1 \\
  \midrule
  DrQA~\citep{chen2017reading}             & -    & -    & -    & 29.8 & - \\
  Multi-Passage BERT~\citep{wang2019multi} & -    & -    & -    & 53.0 & 60.9 \\ 
  Path Retriever~\citep{asai2019learning}  & 31.7 & -    & -    & \textbf{56.5} & \textbf{63.8} \\
  Graph Retriever~\citep{min2019knowledge} & 34.7 & 55.8 & -    & - & -\\
  Hard EM~\citep{min2019discrete}          & 28.8 & 50.9 & -    & - & -\\
  ORQA~\citep{lee2019latent}               & 31.3 & 45.1 & -    & 20.2 & -\\
  REALM~\citep{guu2020realm}               & 40.4 & -    & -    & - & - \\
  DPR~\citep{karpukhin2020dense}           & 41.5 & 57.9 & -    & 36.7 & -\\
  SpanSeqGen~\citep{min2020ambigqa}        & 42.5 & -    & -    & - & -\\
  RAG~\citep{lewis2020retrieval}           & 44.5 & 56.1 & 68.0 & - & -\\
  \midrule
  T5~\citep{roberts2020much} & 36.6 & - & 60.5 & - & -\\
  GPT-3 few shot~\citep{brown2020language}    & 29.9 & - & 71.2 & - & -\\
  \midrule
  Fusion-in-Decoder (base)    & 48.2 & 65.0 & 77.1 & 53.4 & 60.6\\
  Fusion-in-Decoder (large)   & \textbf{51.4} & \textbf{67.6} & \textbf{80.1} & \textbf{56.7} & 63.2 \\
  \bottomrule
\end{tabular}
\caption[Caption]{Comparison to state-of-the-art. On TriviaQA, we report results on the open domain test set (left), and on the hidden test set (right), \url{competitions.codalab.org/competitions/17208\#results}).}
\label{tab:sota}
\end{table*}

\paragraph{Retrieval.}
For the retrieval of support passages, we consider two methods: BM25~\citep{robertson1995okapi} and DPR~\citep{karpukhin2020dense}.
In BM25, passages are represented as bag of words, and the ranking function is based on term and inverse document frequencies.
We use the implementation from Apache Lucene\footnote{\url{lucene.apache.org}} with default parameters, and tokenize questions and passages with SpaCy.\footnote{\url{spacy.io}}
In DPR, passages and questions are represented as dense vector representations, computed using two BERT networks.
The ranking function is the dot product between the query and passage representations. 
Retrieval is performed using approximate nearest neighbors with the FAISS library.\footnote{\url{github.com/facebookresearch/faiss}}

\paragraph{Reading.}
Our generative model for open domain QA is based on a sequence-to-sequence network, pretrained on unsupervised data, such as T5 or BART~\citep{raffel2019exploring,lewis2019bart}.
The model takes as input the question, as well as the support passages, and generates the answer.
More precisely, each retrieved passage and its title are concatenated with the question, and processed independently from other passages by the encoder.
We add special tokens \texttt{question:}, \texttt{title:} and \texttt{context:} before the question, title and text of each passage.
Finally, the decoder performs attention over the concatenation of the resulting representations of all the retrieved passages.
The model thus performs evidence fusion in the decoder only, and we refer to it as \emph{Fusion-in-Decoder}.

By processing passages independently in the encoder, but jointly in the decoder, this method differs from \citet{min2020ambigqa} and \citet{lewis2020retrieval}.
Processing passages independently in the encoder allows to scale to large number of contexts, as it only performs self attention over one context at a time.
This means that the computation time of the model grows linearly with the number of passages, instead of quadratically.
On the other hand, processing passages jointly in the decoder allows to better aggregate evidence from multiple passages.

\section{Experiments}
In this section, we report empirical evaluations of Fusion-in-Decoder for open domain QA.

\begin{figure*}[t]
\begin{center} 
\includegraphics{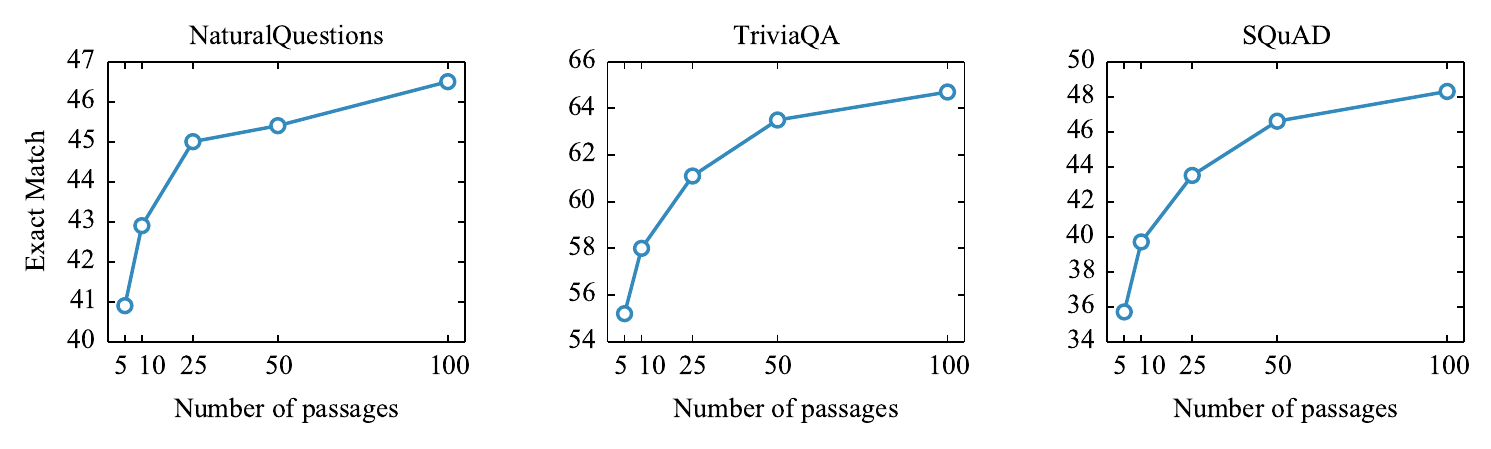}
\caption{Performance of Fusion-in-Decoder (base) on valid sets as a function of the number of retrieved passages.}
\label{fig:contextscaling}
\end{center}
\end{figure*}

\paragraph{Datasets.}
We consider the following datasets, and use the same setting as \citet{lee2019latent}:
\begin{itemize}
\item NaturalQuestions~\citep{kwiatkowski2019NQ} contains questions corresponding to Google search queries. 
The open-domain version of this dataset is obtained by discarding answers with more than 5 tokens.
\item TriviaQA~\citep{JoshiTriviaQA2017} contains questions gathered from trivia and quiz-league websites.
The \emph{unfiltered} version of TriviaQA is used for open-domain question answering.
\item SQuAD v1.1~\citep{rajpurkar2016squad} is a reading comprehension dataset. 
Given a paragraph extracted from Wikipedia, annotators were asked to write questions, for which the answer is a span from the corresponding paragraph.
\end{itemize}
Following~\citet{lee2019latent} we use the validation as test, and keep 10\% of the training set for validation.
We use the Wikipedia dumps from Dec. 20, 2018 for NQ and TriviaQA and from Dec. 21, 2016 for SQuAD.
We apply the same preprocessing as \citet{chen2017reading,karpukhin2020dense}, leading to passages of 100 words, which do not overlap.

\paragraph{Evaluation.}
Predicted answers are evaluated with the standard exact match metric (EM), as introduced by \citet{rajpurkar2016squad}.
A generated answer is considered correct if it matches any answer of the list of acceptable answers after normalization.
This normalization step consists in lowercasing and removing articles, punctuation and duplicated whitespace.

\paragraph{Technical details.}
We initialize our models with the pretrained T5 models~\citep{raffel2019exploring}, available in the HuggingFace Transformers library.\footnote{\url{github.com/huggingface/transformers}}
We consider two model sizes, base and large, containing respectively 220M and 770M parameters.
We fine-tune the models on each dataset independently, using Adam~\citep{kingma2014adam} with a constant learning rate of $10^{-4}$ and a dropout rate of 10\%.
We train the model for 10k gradient steps, with a batch size of 64, using 64 Tesla V100 32Gb.
We evaluate models every 500 steps and select the best one on the validation set based on the Exact Match score.
During training on NaturalQuestions and SQuAD, we sample the target among the list of answers, while for TriviaQA, we use the unique human-generated answer. For TriviaQA, answers in uppercase are normalized by converting all letters in lowercase except the first letter of each word, using the \texttt{title} Python string method.
For both training and testing, we retrieve 100 passages (unless said otherwise), and truncate them to 250 word pieces.
Following the results of~\citet{karpukhin2020dense}, passages are retrieved with DPR for NQ and TriviaQA, and with BM25 for SQuAD.
We generate answers by using greedy decoding.

\paragraph{Comparison to state-of-the-art.}
In table~\ref{tab:sota}, we compare the results obtained by Fusion-in-Decoder with existing approaches for open domain question answering.
We observe that while conceptually simple, this method outperforms existing work on the NaturalQuestion and TriviaQA benchmarks.
%On SQuAD we match the performance of the state-of-the art method based on Exact Match. For the F1 score, we obtain 60.6 with a base model, and 63.2 with a large model, versus 63.8 for Graph Retriever~\citep{min2019knowledge}.
In particular, generative models seem to perform well when evidence from multiple passages need to be aggregated, compared to extractive approaches.
Our method also performs better than other generative models, showing that scaling to large number of passages and processing them jointly leads to improvement in accuracy.
Second, we observe that using additional knowledge in generative models by using retrieval lead to important performance gains.
On NaturalQuestions, the \emph{closed book} T5 model obtains 36.6\% accuracy with 11B parameters, 
while our approach obtains 44.1\% with 770M parameters plus Wikipedia with BM25 retrieval.
Both methods use roughly the same amount of memory to store information, indicating that text based explicit memories are competitive for knowledge retrieval tasks.

%In the next section, we investigate how the Fusion-in-Decoder method scales with respect to the number of retrieved passages.

\paragraph{Scaling with number of passages.}
In Figure~\ref{fig:contextscaling}, we report the performance with respect to the number of retrieved passages.
In particular, we observe that increasing the number of passages from 10 to 100 leads to 6\% improvement on TriviaQA and 3.5\% improvement on NaturalQuestions.
On the other hand, the performance of most extractive models seems to peak around 10 to 20 passages~\citep{wang2019multi,yang2019end}.
We believe that this is evidence that sequence-to-sequence models are good at combining informations from multiple passages.

\paragraph{Impact of the number of training passages.}
In the previous section, the model was trained and evaluated with the same number of passages.
To reduce the training computational budget, a simple solution consists in training the model with fewer passages.
In Table~\ref{tab:passages}, we report the performance obtained by training with different numbers of passages, while testing with 100 passages.
We observe that reducing the number of training passages leads to a decrease of accuracy.
Further, we propose to finetune the previous models using 100 passages for 1000 steps.
This allows to reduce the accuracy gap, while using significantly less computational resources:
we can reach 46.0 EM on NaturalQuestions, using 147 GPU hours, compared to 425 GPU hours when training on 100 passages.

\begin{table*}[t]
\centering
\begin{tabular}{x{4cm}cccc}
  \toprule
  & \multicolumn{2}{c}{NaturalQuestions} & \multicolumn{2}{c}{TriviaQA} \\
  Training Passages & w/o finetuning & w/ finetuning & w/o finetuning & w/ finetuning \\
  \midrule
  5 & 37.8 & 45.0 & 58.1 & 64.2\\
  10 & 42.3 & 45.3 & 61.1 & 63.6\\
  25 & 45.3 & 46.0 & 63.2 & 64.2\\
  50 & 45.7 & 46.0 & 64.2 & 64.3\\
  100 & 46.5 & - & 64.7 & -\\
  \bottomrule
\end{tabular}
\caption[Caption]{Performance depending on the number of passages used during training. Exact Match scores are reported on dev sets.}
\label{tab:passages}
\end{table*}

\section{Conclusion}
In this paper, we study a simple approach to open domain question answering, which relies on retrieving support passages before processing them with a generative model.
We show that while conceptually simple, this approach is competitive with existing methods, and that it scales well with the number of retrieved passages.
In future work, we plan to make this model more efficient, in particular when scaling to large number of support passages.
We also plan to integrate the retrieval in our model, and to learn the whole system end-to-end.

\bibliography{references}

\begin{thebibliography}{34}
\expandafter\ifx\csname natexlab\endcsname\relax\def\natexlab#1{#1}\fi

\bibitem[{Asai et~al.(2020)Asai, Hashimoto, Hajishirzi, Socher, and
  Xiong}]{asai2019learning}
Akari Asai, Kazuma Hashimoto, Hannaneh Hajishirzi, Richard Socher, and Caiming
  Xiong. 2020.
\newblock Learning to retrieve reasoning paths over wikipedia graph for
  question answering.
\newblock In \emph{Proc. ICLR}.

\bibitem[{Brown et~al.(2020)Brown, Mann, Ryder, Subbiah, Kaplan, Dhariwal,
  Neelakantan, Shyam, Sastry, Askell et~al.}]{brown2020language}
Tom~B Brown, Benjamin Mann, Nick Ryder, Melanie Subbiah, Jared Kaplan, Prafulla
  Dhariwal, Arvind Neelakantan, Pranav Shyam, Girish Sastry, Amanda Askell,
  et~al. 2020.
\newblock Language models are few-shot learners.
\newblock \emph{arXiv preprint arXiv:2005.14165}.

\bibitem[{Chen et~al.(2017)Chen, Fisch, Weston, and Bordes}]{chen2017reading}
Danqi Chen, Adam Fisch, Jason Weston, and Antoine Bordes. 2017.
\newblock Reading {W}ikipedia to answer open-domain questions.
\newblock In \emph{Proc. ACL}.

\bibitem[{Clark and Gardner(2018)}]{clark2017simple}
Christopher Clark and Matt Gardner. 2018.
\newblock Simple and effective multi-paragraph reading comprehension.
\newblock In \emph{Proc. ACL}.

\bibitem[{Devlin et~al.(2019)Devlin, Chang, Lee, and
  Toutanova}]{devlin2018bert}
Jacob Devlin, Ming-Wei Chang, Kenton Lee, and Kristina Toutanova. 2019.
\newblock {BERT}: Pre-training of deep bidirectional transformers for language
  understanding.
\newblock In \emph{Proc. NAACL}.

\bibitem[{Fan et~al.(2019)Fan, Jernite, Perez, Grangier, Weston, and
  Auli}]{fan2019eli5}
Angela Fan, Yacine Jernite, Ethan Perez, David Grangier, Jason Weston, and
  Michael Auli. 2019.
\newblock {ELI}5: Long form question answering.
\newblock In \emph{Proc. ACL}.

\bibitem[{Guu et~al.(2020)Guu, Lee, Tung, Pasupat, and Chang}]{guu2020realm}
Kelvin Guu, Kenton Lee, Zora Tung, Panupong Pasupat, and Ming-Wei Chang. 2020.
\newblock Realm: Retrieval-augmented language model pre-training.
\newblock \emph{arXiv preprint arXiv:2002.08909}.

\bibitem[{Jiang et~al.(2019)Jiang, Xu, Araki, and Neubig}]{jiang2019can}
Zhengbao Jiang, Frank~F Xu, Jun Araki, and Graham Neubig. 2019.
\newblock How can we know what language models know?
\newblock \emph{arXiv preprint arXiv:1911.12543}.

\bibitem[{Joshi et~al.(2017)Joshi, Choi, Weld, and
  Zettlemoyer}]{JoshiTriviaQA2017}
Mandar Joshi, Eunsol Choi, Daniel~S. Weld, and Luke Zettlemoyer. 2017.
\newblock Triviaqa: A large scale distantly supervised challenge dataset for
  reading comprehension.
\newblock In \emph{Proc. ACL}.

\bibitem[{Karpukhin et~al.(2020)Karpukhin, O{\u{g}}uz, Min, Wu, Edunov, Chen,
  and Yih}]{karpukhin2020dense}
Vladimir Karpukhin, Barlas O{\u{g}}uz, Sewon Min, Ledell Wu, Sergey Edunov,
  Danqi Chen, and Wen-tau Yih. 2020.
\newblock Dense passage retrieval for open-domain question answering.
\newblock \emph{arXiv preprint arXiv:2004.04906}.

\bibitem[{Kingma and Ba(2014)}]{kingma2014adam}
Diederik~P Kingma and Jimmy Ba. 2014.
\newblock Adam: A method for stochastic optimization.
\newblock \emph{arXiv preprint arXiv:1412.6980}.

\bibitem[{Ko{\v{c}}isk{\`y} et~al.(2018)Ko{\v{c}}isk{\`y}, Schwarz, Blunsom,
  Dyer, Hermann, Melis, and Grefenstette}]{kovcisky2018narrativeqa}
Tom{\'a}{\v{s}} Ko{\v{c}}isk{\`y}, Jonathan Schwarz, Phil Blunsom, Chris Dyer,
  Karl~Moritz Hermann, G{\'a}bor Melis, and Edward Grefenstette. 2018.
\newblock The {N}arrative{QA} reading comprehension challenge.
\newblock \emph{TACL}.

\bibitem[{Kwiatkowski et~al.(2019)Kwiatkowski, Palomaki, Redfield, Collins,
  Parikh, Alberti, Epstein, Polosukhin, Kelcey, Devlin, Lee, Toutanova, Jones,
  Chang, Dai, Uszkoreit, Le, and Petrov}]{kwiatkowski2019NQ}
Tom Kwiatkowski, Jennimaria Palomaki, Olivia Redfield, Michael Collins, Ankur
  Parikh, Chris Alberti, Danielle Epstein, Illia Polosukhin, Matthew Kelcey,
  Jacob Devlin, Kenton Lee, Kristina~N. Toutanova, Llion Jones, Ming-Wei Chang,
  Andrew Dai, Jakob Uszkoreit, Quoc Le, and Slav Petrov. 2019.
\newblock Natural {Q}uestions: a benchmark for question answering research.
\newblock \emph{TACL}.

\bibitem[{Lee et~al.(2018)Lee, Yun, Kim, Ko, and Kang}]{lee2018ranking}
Jinhyuk Lee, Seongjun Yun, Hyunjae Kim, Miyoung Ko, and Jaewoo Kang. 2018.
\newblock Ranking paragraphs for improving answer recall in open-domain
  question answering.
\newblock In \emph{Proc. EMNLP}.

\bibitem[{Lee et~al.(2019)Lee, Chang, and Toutanova}]{lee2019latent}
Kenton Lee, Ming-Wei Chang, and Kristina Toutanova. 2019.
\newblock Latent retrieval for weakly supervised open domain question
  answering.
\newblock In \emph{Proc. ACL}.

\bibitem[{Lewis et~al.(2019)Lewis, Liu, Goyal, Ghazvininejad, Mohamed, Levy,
  Stoyanov, and Zettlemoyer}]{lewis2019bart}
Mike Lewis, Yinhan Liu, Naman Goyal, Marjan Ghazvininejad, Abdelrahman Mohamed,
  Omer Levy, Ves Stoyanov, and Luke Zettlemoyer. 2019.
\newblock {B}{A}{R}{T}: Denoising sequence-to-sequence pre-training for natural
  language generation, translation, and comprehension.
\newblock \emph{arXiv preprint arXiv:1910.13461}.

\bibitem[{Lewis et~al.(2020)Lewis, Perez, Piktus, Petroni, Karpukhin, Goyal,
  K{\"u}ttler, Lewis, Yih, Rockt{\"a}schel et~al.}]{lewis2020retrieval}
Patrick Lewis, Ethan Perez, Aleksandara Piktus, Fabio Petroni, Vladimir
  Karpukhin, Naman Goyal, Heinrich K{\"u}ttler, Mike Lewis, Wen-tau Yih, Tim
  Rockt{\"a}schel, et~al. 2020.
\newblock Retrieval-augmented generation for knowledge-intensive nlp tasks.
\newblock \emph{arXiv preprint arXiv:2005.11401}.

\bibitem[{Min et~al.(2019{\natexlab{a}})Min, Chen, Hajishirzi, and
  Zettlemoyer}]{min2019discrete}
Sewon Min, Danqi Chen, Hannaneh Hajishirzi, and Luke Zettlemoyer.
  2019{\natexlab{a}}.
\newblock A discrete hard {EM} approach for weakly supervised question
  answering.
\newblock In \emph{Proc. EMNLP-IJCNLP}.

\bibitem[{Min et~al.(2019{\natexlab{b}})Min, Chen, Zettlemoyer, and
  Hajishirzi}]{min2019knowledge}
Sewon Min, Danqi Chen, Luke Zettlemoyer, and Hannaneh Hajishirzi.
  2019{\natexlab{b}}.
\newblock Knowledge guided text retrieval and reading for open domain question
  answering.
\newblock \emph{arXiv preprint arXiv:1911.03868}.

\bibitem[{Min et~al.(2020)Min, Michael, Hajishirzi, and
  Zettlemoyer}]{min2020ambigqa}
Sewon Min, Julian Michael, Hannaneh Hajishirzi, and Luke Zettlemoyer. 2020.
\newblock Ambigqa: Answering ambiguous open-domain questions.
\newblock \emph{arXiv preprint arXiv:2004.10645}.

\bibitem[{Peters et~al.(2018)Peters, Neumann, Iyyer, Gardner, Clark, Lee, and
  Zettlemoyer}]{peters2018deep}
Matthew Peters, Mark Neumann, Mohit Iyyer, Matt Gardner, Christopher Clark,
  Kenton Lee, and Luke Zettlemoyer. 2018.
\newblock Deep contextualized word representations.
\newblock In \emph{Proc. NAACL}.

\bibitem[{Petroni et~al.(2019)Petroni, Rockt{\"a}schel, Riedel, Lewis, Bakhtin,
  Wu, and Miller}]{petroni2019language}
Fabio Petroni, Tim Rockt{\"a}schel, Sebastian Riedel, Patrick Lewis, Anton
  Bakhtin, Yuxiang Wu, and Alexander Miller. 2019.
\newblock Language models as knowledge bases?
\newblock In \emph{Proc. EMNLP-IJCNLP}.

\bibitem[{Radford et~al.(2019)Radford, Wu, Child, Luan, Amodei, and
  Sutskever}]{radford2019language}
Alec Radford, Jeffrey Wu, Rewon Child, David Luan, Dario Amodei, and Ilya
  Sutskever. 2019.
\newblock Language models are unsupervised multitask learners.
\newblock \emph{OpenAI Technical Report}.

\bibitem[{Raffel et~al.(2019)Raffel, Shazeer, Roberts, Lee, Narang, Matena,
  Zhou, Li, and Liu}]{raffel2019exploring}
Colin Raffel, Noam Shazeer, Adam Roberts, Katherine Lee, Sharan Narang, Michael
  Matena, Yanqi Zhou, Wei Li, and Peter~J Liu. 2019.
\newblock Exploring the limits of transfer learning with a unified text-to-text
  transformer.
\newblock \emph{arXiv preprint arXiv:1910.10683}.

\bibitem[{Rajpurkar et~al.(2016)Rajpurkar, Zhang, Lopyrev, and
  Liang}]{rajpurkar2016squad}
Pranav Rajpurkar, Jian Zhang, Konstantin Lopyrev, and Percy Liang. 2016.
\newblock {SQ}u{AD}: 100,000+ questions for machine comprehension of text.
\newblock In \emph{Proc. EMNLP}.

\bibitem[{Reddy et~al.(2019)Reddy, Chen, and Manning}]{reddy2019coqa}
Siva Reddy, Danqi Chen, and Christopher~D Manning. 2019.
\newblock {CoQA}: A conversational question answering challenge.
\newblock \emph{TACL}.

\bibitem[{Roberts et~al.(2020)Roberts, Raffel, and Shazeer}]{roberts2020much}
Adam Roberts, Colin Raffel, and Noam Shazeer. 2020.
\newblock How much knowledge can you pack into the parameters of a language
  model?
\newblock \emph{arXiv preprint arXiv:2002.08910}.

\bibitem[{Robertson et~al.(1995)Robertson, Walker, Jones, Hancock-Beaulieu,
  Gatford et~al.}]{robertson1995okapi}
Stephen~E Robertson, Steve Walker, Susan Jones, Micheline~M Hancock-Beaulieu,
  Mike Gatford, et~al. 1995.
\newblock Okapi at {TREC}-3.
\newblock \emph{NIST Special Publication Sp}.

\bibitem[{Talmor et~al.(2019)Talmor, Elazar, Goldberg, and
  Berant}]{talmor2019olmpics}
Alon Talmor, Yanai Elazar, Yoav Goldberg, and Jonathan Berant. 2019.
\newblock o{LM}pics--on what language model pre-training captures.
\newblock \emph{arXiv preprint arXiv:1912.13283}.

\bibitem[{Voorhees et~al.(1999)}]{voorhees1999trec}
Ellen~M Voorhees et~al. 1999.
\newblock The {TREC}-8 question answering track report.
\newblock In \emph{TREC}.

\bibitem[{Wang et~al.(2018{\natexlab{a}})Wang, Yu, Guo, Wang, Klinger, Zhang,
  Chang, Tesauro, Zhou, and Jiang}]{wang2018r}
Shuohang Wang, Mo~Yu, Xiaoxiao Guo, Zhiguo Wang, Tim Klinger, Wei Zhang, Shiyu
  Chang, Gerry Tesauro, Bowen Zhou, and Jing Jiang. 2018{\natexlab{a}}.
\newblock R$^3$: Reinforced ranker-reader for open-domain question answering.
\newblock In \emph{Proc. AAAI}.

\bibitem[{Wang et~al.(2018{\natexlab{b}})Wang, Yu, Jiang, Zhang, Guo, Chang,
  Wang, Klinger, Tesauro, and Campbell}]{wang2017evidence}
Shuohang Wang, Mo~Yu, Jing Jiang, Wei Zhang, Xiaoxiao Guo, Shiyu Chang, Zhiguo
  Wang, Tim Klinger, Gerald Tesauro, and Murray Campbell. 2018{\natexlab{b}}.
\newblock Evidence aggregation for answer re-ranking in open-domain question
  answering.
\newblock In \emph{Proc. ICLR}.

\bibitem[{Wang et~al.(2019)Wang, Ng, Ma, Nallapati, and Xiang}]{wang2019multi}
Zhiguo Wang, Patrick Ng, Xiaofei Ma, Ramesh Nallapati, and Bing Xiang. 2019.
\newblock Multi-passage {BERT}: A globally normalized {BERT} model for
  open-domain question answering.
\newblock In \emph{Proc. EMNLP-IJCNLP}.

\bibitem[{Yang et~al.(2019)Yang, Xie, Lin, Li, Tan, Xiong, Li, and
  Lin}]{yang2019end}
Wei Yang, Yuqing Xie, Aileen Lin, Xingyu Li, Luchen Tan, Kun Xiong, Ming Li,
  and Jimmy Lin. 2019.
\newblock End-to-end open-domain question answering with {BERT}serini.
\newblock In \emph{Proc. NAACL (Demonstrations)}.

\end{thebibliography}
\bibliographystyle{acl_natbib}

%\clearpage
%\input{appendix}
%\newpage

%\input{full_results}
%\vspace{10cm}

\end{document}